\documentclass[letterpaper]{article} 
\usepackage{aaai25}  
\usepackage{times}  
\usepackage{helvet}  
\usepackage{courier}  
\usepackage[hyphens]{url}  
\usepackage{graphicx} 
\usepackage{natbib}  
\usepackage{caption} 
\usepackage{algorithm}
\usepackage{algorithmic}
\usepackage{color}
\usepackage{mathrsfs}
\usepackage{amsmath}
\usepackage{amsfonts}
\usepackage{amssymb}
\usepackage{multirow} 
\usepackage{subfigure} 
\usepackage{booktabs}

\usepackage{newfloat}
\usepackage{listings}
\DeclareCaptionStyle{ruled}{labelfont=normalfont,labelsep=colon,strut=off} 
\floatstyle{ruled}
\newfloat{listing}{tb}{lst}{}
\floatname{listing}{Listing}
\title{Mitigating Hallucinations in Large Vision-Language Models by Adaptively Constraining Information Flow}
\author{
    Jiaqi Bai\textsuperscript{\rm 1,\rm2},
    Hongcheng Guo\textsuperscript{\rm 3},
    Zhongyuan Peng\textsuperscript{\rm 4},
    Jian Yang\textsuperscript{\rm 3},\\
    Zhoujun Li\textsuperscript{\rm 3},
    Mohan Li\textsuperscript{\rm 1,\rm2}\thanks{Corresponding author.},
    Zhihong Tian\textsuperscript{\rm 1,\rm2}
}
\affiliations{
    \textsuperscript{\rm 1}Cyberspace Institute of Advanced Technology, Guangzhou University, China \\
    \textsuperscript{\rm 2}Huangpu Research School of Guangzhou University, China \\
    \textsuperscript{\rm 3}CCSE, Beihang University, China \\
    \textsuperscript{\rm 4}University of the Chinese Academy of Sciences, China \\
    \{jqbai, limohan, tianzhihong\}@gzhu.edu.cn, 
    \{hongchengguo, jiaya, lizj\}@buaa.edu.cn, \\
    zhongyuan.peng@cripac.ia.ac.cn
    
}

\usepackage{bibentry}

\begin{document}

\maketitle

\begin{abstract}

Large vision-language models show tremendous potential in understanding visual information through human languages.
However, they are prone to suffer from object hallucination, i.e., the generated image descriptions contain objects that do not exist in the image.
In this paper, we reveal that object hallucination can be attributed to overconfidence in irrelevant visual features when soft visual tokens map to the LLM's word embedding space.
Specifically, by figuring out the semantic similarity between visual tokens and LLM's word embedding, we observe that the smoothness of similarity distribution strongly correlates with the emergence of object hallucinations.
To mitigate hallucinations, we propose using the Variational Information Bottleneck (VIB) to alleviate overconfidence by introducing stochastic noise, facilitating the constraining of irrelevant information.
Furthermore, we propose an entropy-based noise-controlling strategy to enable the injected noise to be adaptively constrained regarding the smoothness of the similarity distribution.
We adapt the proposed \textsc{AdaVIB} across distinct model architectures.
Experimental results demonstrate that the proposed \textsc{AdaVIB} mitigates object hallucinations by effectively alleviating the overconfidence in irrelevant visual features, with consistent improvements on two object hallucination benchmarks.

\end{abstract}

\begin{links}
\link{Code}{https://github.com/jiaqi5598/AdaVIB}
\end{links}

\section{Introduction}

\begin{figure}[ht]
\centering
\includegraphics[width=8.cm]{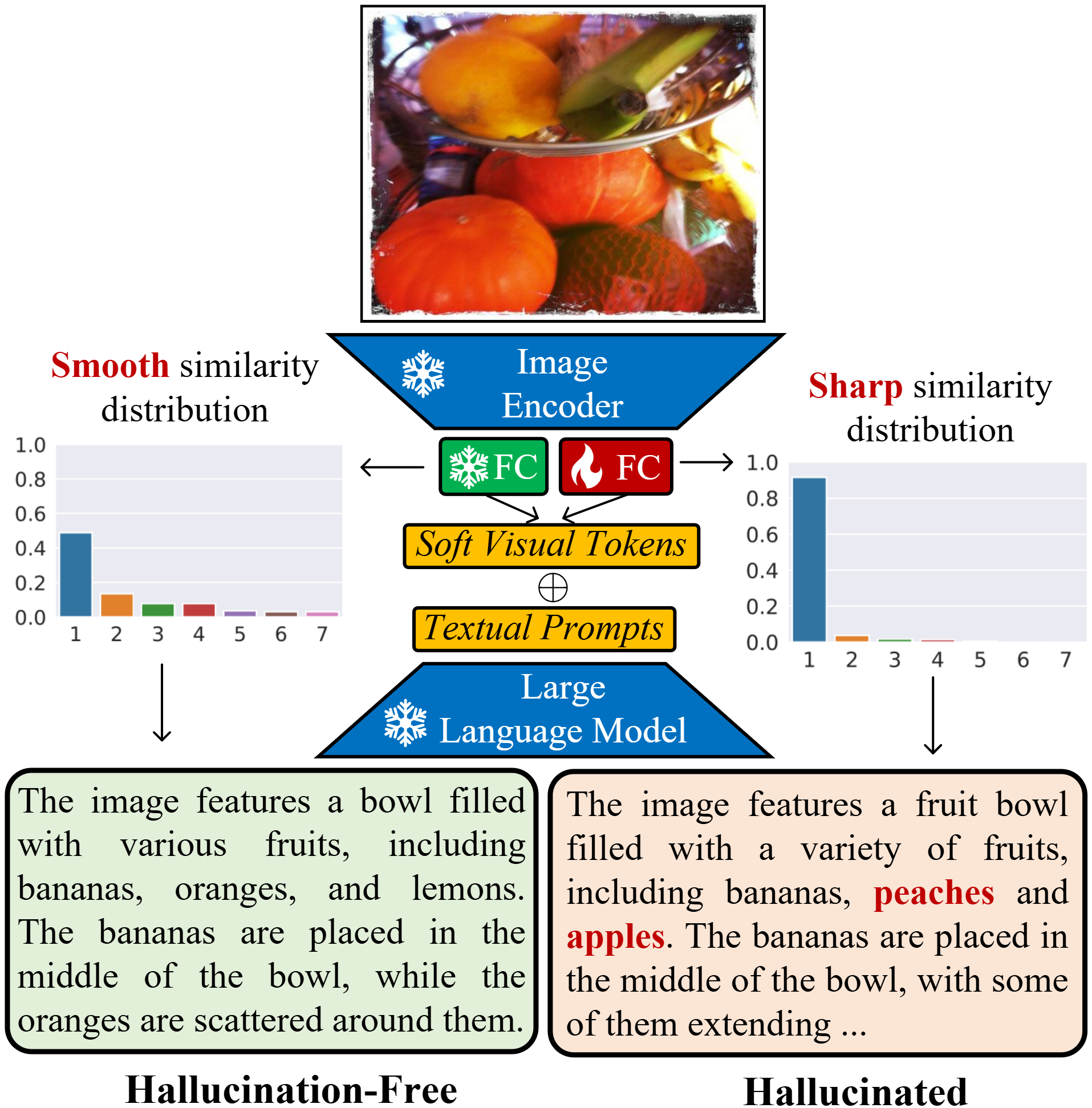}
\caption{Impact on the smoothness of the similarity distribution correlates with the emergence of object hallucinations. We use the normalized dot product to measure the semantic similarity between soft visual tokens and LLM's word embedding. 
The y-axis of the similarity distribution denotes the similarity score. The x-axis is the top-ranked LLM's token sorting in descending order.}
\label{intro_case}
\end{figure}

Large Vision-Language Models (LVLMs)~\cite{zhu2023minigpt,dai2023instructblip,bai2023qwen} have recently attracted increasing attention.
Due to the superiority of generating contextually relevant natural language descriptions grounded on visual patterns, LVLMs have shown impressive performance on various vision-language tasks, including image captioning~\cite{deng2009imagenet,lin2014microsoft}, visual question answering~\cite{antol2015vqa,li2023evaluating} and multimodal machine translation~\cite{yao2020multimodal,guo2022lvp}.
Despite their success, LVLMs are prone to object hallucinations~\cite{rohrbach2018object,biten2022let}, where the generated image descriptions are inconsistent with the objects that appear in the grounded image.
This inconsistency has significantly affected the reliability and applicability of LVLMs, especially in scenarios that demand precise judgment.

Most existing works focus on mitigating object hallucinations by reducing over-reliance on prior knowledge of LLMs through the decoding process~\cite{huang2024opera,favero2024multi}.
They devise various methods to penalize specific patterns that may induce hallucinations~\cite{huang2024opera,leng2024mitigating} or curate a training dataset to address statistical biases arising from co-occurrence and spurious correlation among objects~\cite{biten2022let,zhou2024analyzing,liu2023mitigating}. 
Despite their efforts, most of these works rarely focus on narrowing the modality gap between visual patterns and textual descriptions.
Recently, researchers have revealed that the vision-language connector plays a significant role in mitigating object hallucinations of LVLMs~\cite{li2023blip,sun2023aligning,jiang2024hallucination}. 
On the one hand, despite their impressive performance on various visual understanding tasks, existing vision encoding techniques are still challenging in expressing visual patterns precisely~\cite{cho2022fine,li2024monkey,jain2024vcoder}.
Hence, compressing irrelevant visual features encoded by the vision encoder is essential to generate hallucination-free descriptions.
On the other hand, although previous work investigated various modality alignment approaches, such as Q-Formers~\cite{li2023blip,Dai2023InstructBLIPTG} and lightweight projectors~\cite{DBLP:journals/patterns/LiuLTLZ22,alayrac2022flamingo,gao2023llama,chen2023minigpt,liu2024improved}, the encoded soft visual tokens are still far from ideal in compressing irrelevant while preserving relevant 
visual information faithful to input images.

In this paper, we reveal that object hallucinations can be attributed to \emph{overconfidence} in irrelevant visual features when soft visual tokens project to the word embedding space of LLM.
As Figure \ref{intro_case} shows, we present two distinct cases illustrating how the smoothness of the similarity distribution between soft visual tokens\footnote{We apply an average pooling over all visual tokens to obtain an overall representation.} and the LLM's word embedding correlates with the emergence of object hallucinations. 
Both frozen and fine-tuned fully connected (FC) layers, referred to as the \emph{vision-language projector}, are introduced for comparison to project encoded visual features into soft visual tokens.
By examining the normalized dot product similarity between soft visual tokens and the word embedding of LLM, we observe that the LVLMs equipped with a frozen FC layer (the green block on the left) generate a hallucination-free description with a smoother similarity distribution.
In contrast, the variant equipped with a fine-tuned FC layer (the red block on the right) generates a hallucinated description with a sharper similarity distribution.
This phenomenon indicates that the smoothness of the similarity distribution strongly correlates with the emergence of object hallucinations.
A sharp similarity distribution indicates the occurrence of the \emph{overconfidence} problem, which results from overfitting on statistically spurious correlations with the irrelevant visual features during training.

Therefore, alleviating the \emph{overconfidence} on irrelevant features when soft visual tokens mapping to the LLM's word embedding is essential to mitigate object hallucinations, and it is critical to constrain information flow to soft visual tokens by elaborately devising the vision-language projector.

Motivated by the above analysis, we propose \textsc{AdaVIB}, a lightweight fine-tuning method that uses Variational Information Bottleneck (VIB)~\cite{alemi2016deep} to mitigate object hallucinations.
The proposed \textsc{AdaVIB} only requires fine-tuning weights of the vision-language projector with other modules frozen.
It mitigates \emph{overconfidence} problem by introducing a compression term to regularize the training of vision-language projector.
Introducing the compression term can be regarded as adding stochastic noise to soft visual tokens during training, and increasing this noise constrains the information flow to them, which decreases the \emph{overconfidence} in irrelevant visual features when visual tokens mapping to the LLM's word embedding space.
To adaptively constrain the visual information flow to the soft visual tokens, we propose an entropy-based noise-controlling mechanism. 
The proposal adaptively constrains the injected noise regarding the smoothness of the similarity distribution between soft visual tokens and the LLM's word embedding, capturing the dynamic nature of a specific sample.
We adapt \textsc{AdaVIB} on two distinct LVLM architectures, including MiniGPT4~\cite{zhu2023minigpt} and LLava-1.5~\cite{liu2024improved}.
Experimental results demonstrate the effectiveness of \textsc{AdaVIB} in mitigating the overconfidence problem by effectively smoothing the similarity distribution between soft visual tokens and LLM's word embedding, with consistent improvements on two object hallucination benchmarks. 

Our contributions are three-fold:
\textbf{i):} We are the first to use VIB to mitigate object hallucinations, which decreases the overconfidence in irrelevant visual features when soft visual tokens map to the word embedding of LLM.
\textbf{ii):} We propose \textsc{AdaVIB}, an entropy-based noise controlling strategy to adaptive constrain the information conveyed by visual tokens, regarding the smoothness of similarity distribution to LLM's word embedding.
\textbf{iii):} Comprehensive experiments demonstrate the effectiveness of \textsc{AdaVIB} in mitigating object hallucinations. The proposed approach yields consistent improvements on two object hallucination benchmarks across different model architectures.

\section{Related Work}

\subsection{Information Bottleneck}

The Information Bottleneck (IB) principle~\cite{tishby2000information,fischer2020conditional} is an excellent concept for regularizing internal representations to minimize the mutual information by compressing the original input representation. 
The compressed representation can further improve the model's generalization capability by ignoring irrelevant features in the original input.
IB has been widely adopted for many machine learning tasks~\cite{peng2018variational,li2019specializing,belinkov2020variational}, such as image generation~\cite{peng2018variational}, explanation regeneration~\cite{li2023explanation}, retrieval-augmented generation~\cite{zhu2024information}, and so on.

Based on the IB principle, \citet{alemi2016deep} introduced variational information bottleneck (VIB), a variational approach that can be instantiated in deep neural networks, inspired by a similar approach in variational autoencoders (VAE)~\cite{kingma2013auto}.
It has been applied in the study of parsing~\cite{li2019specializing}, natural language inference~\cite{belinkov2020variational}, and graph structure learning~\cite{sun2022graph}.
Compared to the above work, our work, to the best of our knowledge, is the first attempt to investigate VIB as a regularization technique to mitigate object hallucinations in LVLMs. 

\subsection{Object Hallucinations in LVLMs}

Mitigating object hallucinations~\cite{rohrbach2018object,biten2022let} has been a long-standing challenge in realizing a trustworthy AI system.
This issue can be attributed to several possible reasons, e.g., insufficient understanding of the real-world knowledge~\cite{leng2024mitigating,huang2024opera}, statistical bias in training data~\cite{tang2021codes,biten2022let,zhou2024analyzing} or uncertainty to the objects present in the image~\cite{cho2022fine,DBLP:journals/patterns/LiuLTLZ22,li2024monkey}.

To mitigate object hallucinations, existing studies typically involved methods such as contrastive decoding~\cite{leng2024mitigating,huang2024opera}, balance co-occurrence patterns through data augmentation~\cite{zhou2024analyzing,liu2024improved} and devise an aligner to narrow the modality gap between vision and languages~\cite{zhu2023minigpt,dai2023instructblip}.
For example, \citet{chuang2023dola} and \citet{leng2024mitigating} introduced distinct decoding approaches to estimate output distributions by contrasting different sources.
They effectively alleviate object hallucinations by reducing the over-reliance on the prior knowledge of a single source.
Some studies \cite{biten2022let,zhou2024analyzing} augmented training data by analyzing key factors underlying object hallucination, effectively reducing the statistical bias to alleviate object hallucinations.
Compared to the above work, our work starts with an observation that the object hallucinations stem from the overconfidence problem. 
Based on this, we propose a VIB-based approach to mitigate object hallucinations by alleviating such a problem.

\section{Methodology}

\begin{figure*}[t]
\centering
\includegraphics[width=15.8cm]{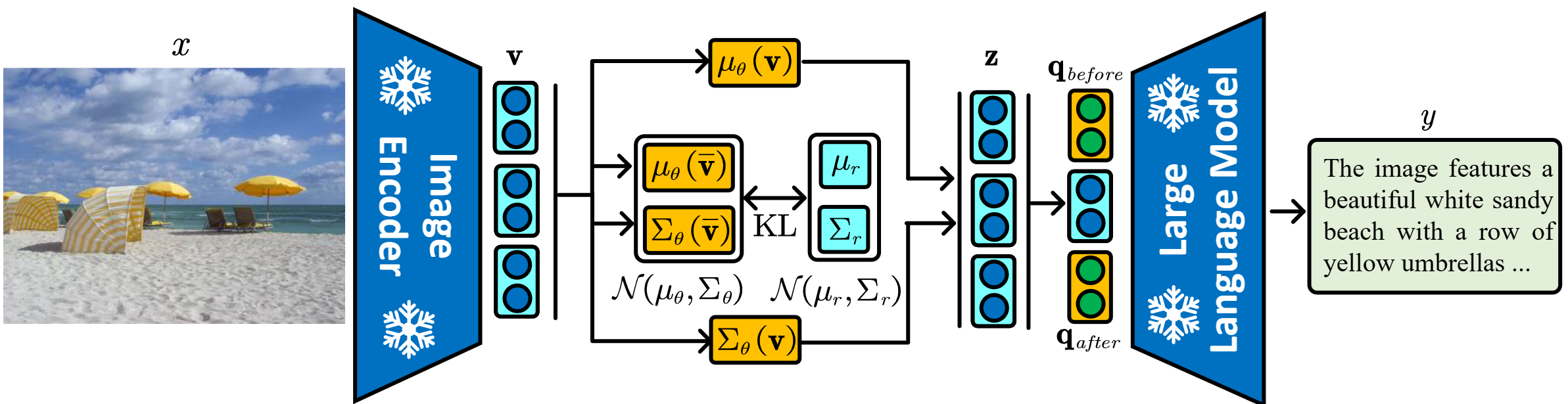}
\caption{The model architecture of \textsc{AdaVIB}. \textsc{AdaVIB} compresses the input representations $\mathbf{v}$ into soft visual tokens $\mathbf{z}$ with mean $\mu _{\theta}(\mathbf{v})$ and constrain the irrelevant information by injecting the Gaussian noise with variance $\Sigma _{\theta}(\mathbf{v})$.}
\label{model_arch}
\end{figure*}

\subsection{Problem Formulation}

We consider a general image-to-text generation problem with a dataset $\mathcal{D}=\{(x_i,q_i,y_i)\}_{i=1}^{N}$, where the $i$-th triple $( x_i,q_i,y_i )$ consists of an image $x_i$, an textual prompt $q_i$ and a image description $y_i$.
The goal is to learn a probability distribution $p(y|x,q)$ with $\mathcal{D}$, and thus given a new pair $(x, q)$ in an image-to-text generation, one can generate a response $y$ token-by-token in terms of $p(y |x, q)$. 
The optimization of $p(y |x, q)$ can be formalized by maximizing the following conditional probability:
\begin{equation}
p(y|x,q)=\prod_{t=1}^{| \mathbf{y} |}{p( y_t |y_{<t},x,q )}
\end{equation}

\subsection{Large Vision-Language Models}

Given an input image $x_j$, a pre-trained visual encoder (e.g., CLIP~\cite{radford2021learning}) first encodes $x_j$ into a dense visual representation $\mathbf{x}_j$, which is then fed to a vision-language connector to obtain an intermediate representation $\mathbf{v}_j$, denoted as $\mathbf{v}_j=g_{\theta}( x_j )$, where $\theta$ is a set of parameters.
The vision language connector is optional, and can be arbitrary architectures, such as a Q-Former~\cite{li2023blip,Dai2023InstructBLIPTG}. 
After that, a vision-language projector receives $\mathbf{v}_j$ and maps it to the word embedding space of LLM, yielding a sequence of visual tokens $\mathbf{z}_j$, the vision-language projector can be a Multilayer Perceptron (i.e., MLP):
\begin{equation}
\mathbf{z}_j=\mathbf{W}_h\mathrm{GeLU}\left( \mathbf{W}_z\mathbf{v}_j \right) 
\end{equation}

\noindent where $\mathbf{W}_z$ and $\mathbf{W}_h$ are trainable parameters.
Lastly, by embedding the textual prompt $q$ into embedded representations $\mathbf{q}$, the gathered visual tokens $\mathbf{z}$ are concatenated with $\mathbf{q}$, yielding outputs $y$ token-by-token. 
Suppose $p_{\phi}( y |\mathbf{v},q )$ is an approximation of $p( y|x,q)$ parameterized by $\phi$.
The above process can be optimized by minimizing the following loss:
\begin{equation}
\label{convention_ce}
\min \mathcal{L}_{CE}=\mathbb{E}_{\left( x,q,y \right) \sim \mathcal{D},\mathbf{v}\sim g_{\theta}\left( x \right)}\left[ -\log \left( p_{\phi}\left( \left. y \right|\mathbf{v},q \right) \right) \right] 
\end{equation}

In this paper, we focus on optimizing the vision-language projector with other modules frozen.
Our approach can be easily extended to LVLMs with and without a vision-language connector (e.g., Q-Former) by simply replacing the vision-language projector with our component.
For the convenience of description, we introduce our method upon the structure without the vision-language connector (e.g., LLaVa~\cite{liu2024visual,liu2024improved}), and the inputs of the vision-language projector denote as $\mathbf{v}$ unless explicitly specified.

\subsection{Adaptive Variational Information Bottleneck}

\paragraph{Information Bottleneck} 
The Information Bottleneck (IB)~\cite{tishby2015deep} has 
been demonstrated as a promising principle to find a compression $\mathbf{z}$ for the original input representation $\mathbf{v}$, that maximally compressing irrelevant information to $\mathbf{v}$ while maintaining relevant information to $\mathbf{y}$.
The objective of IB is to minimize the combination of compression loss and prediction loss, which is formalized as follows:
\begin{equation}
\label{eq_ib}
\min \mathcal{L}_{\mathrm{IB}}=\underset{\mathrm{Compression}}{\underbrace{\beta I(\mathbf{v};\mathbf{z})}}-\underset{\mathrm{Prediction}}{\underbrace{I(\mathbf{z};\mathbf{y})}}
\end{equation}

\noindent where $\beta \geqslant 0$ is the Lagrange multiplier for balancing the compression and prediction terms. 
$I( \cdot ;\cdot )$ is the mutual information. 
The former term of Equation \ref{eq_ib} improves the conciseness of the input signal $\mathbf{v}$ by minimizing the inclusion of irrelevant information, encouraging the network to concentrate more on useful content.
The latter enables the network to selectively maintain useful information for supporting the predicted content $\mathbf{y}$ faithful to the input $\mathbf{v}$.

\paragraph{Variational Information Bottleneck}

\citet{alemi2016deep} proposed Variational Information Bottleneck (VIB), a variational estimation of IB by approximating the probability distribution via a neural network:
\begin{equation}
\label{eq_vib}
\begin{aligned}
\min \mathcal{L}_{\mathrm{VIB}}
& = \beta \mathop {\mathbb{E}} \limits_{\mathbf{v}}[ \mathrm{KL}( p_{\theta}( \mathbf{z} |\mathbf{v} ) \| r( \mathbf{z} ) ) ] \\
& + \mathop {\mathbb{E}} \limits_{\mathbf{z}\thicksim p_{\theta}(\mathbf{z}|\mathbf{v})}[-\log p_{\phi}(y|\mathbf{z},q)]
\end{aligned}
\end{equation}

\noindent where $p_{\phi}(y|\mathbf{z},q)$ is an estimation of response $y$ parameterized by $\phi$, given compressed input representation $\mathbf{z}$ and textual prompt $q$.
$r( \mathbf{z})$ and $p_{\theta}( \mathbf{z} |\mathbf{v})$ are the estimation of prior and posterior probability to $\mathbf{z}$, respectively.
The former part of Equation \ref{eq_vib} serves as a compression term, which provides an explicit way to compress input representation $\mathbf{v}$ into $\mathbf{z}$.
The compressed process can be regarded as introducing stochastic noise during training, where the noise can be injected by sampling $\mathbf{z}$ from $p_{\theta}(\mathbf{z}|\mathbf{v})$.
Increasing this noise decreases the information conveyed by $\mathbf{z}$.
The later part of Equation \ref{eq_vib} is a prediction term, which preserves the useful information for the prediction of $y$. 
When $\beta=0$, there is no incentive to inject a noise perturbation; thereby, the VIB degrades to deterministic dimensionality reduction with an MLP, similar to the implementation of Equation \ref{convention_ce}.
During testing, we use the expected value of $\mathbf{z}$ to predict answer $y$ through $p_{\phi}(y|\mathbf{z},q)$.

In our experiment, the prior distribution $r(\mathbf{z})$ and posterior distribution $p_{\theta}(\mathbf{z}|\mathbf{v})$ are modeled by parametric Gaussian distributions, formalized by:
\begin{gather}
r\left( \mathbf{z} \right) =\mathcal{N}\left( \mu _r,\Sigma _r \right) 
\\
p_{\theta}\left( \left. \mathbf{z} \right|\mathbf{v} \right) =\mathcal{N}\left( \mu _{\theta}\left( \mathbf{\bar{v}} \right) ,\Sigma _{\theta}\left( \mathbf{\bar{v}} \right) \right) 
\end{gather}

\noindent where $\mu _r$ and $\mu _{\theta}$ are mean vectors, $\Sigma _r$ and $\Sigma _{\theta}$ are diagonal covariance matrices. 
$\mathbf{\bar{v}}$ is the average-pooling of $\mathbf{v}$ over all tokens.
Because each dimension of these variables is independent and identically distributed, the Kullback-Leibler (KL) divergence of the multivariate Gaussian distribution can be estimated as follows:
\begin{equation}
\begin{aligned}
\label{kl_loss}
\mathrm{KL(}\mathcal{N}_r||\mathcal{N}_{\theta})
& =\frac{1}{2}[\log \frac{\det\mathrm{(}\Sigma _{\theta})}{\det\mathrm{(}\Sigma _r)}-d_{\mathbf{z}}+\mathrm{tr(}\Sigma _{\mathrm{\theta}}^{-1}\Sigma _r)
\\
& +(\mu _{\mathrm{\theta}}-\mu _r)^{\top}\Sigma _{\mathrm{\theta}}^{-1}(\mu _{\mathrm{\theta}}-\mu _r)]
\end{aligned}
\end{equation}

\noindent where $d_{\mathbf{z}}$ is the dimensionality of $\mathbf{z}$. 
Similar to \citet{li2019specializing}, we apply the reparameterization strategy \cite{kingma2013auto} to approximate gradients backpropagate to $\mathbf{v}$, formalized by: 
\begin{equation}
\label{repara_trick}
\mathbf{z}=\mu _{\theta}( \mathbf{v} ) +\Sigma _{\theta}( \mathbf{v} ) \odot \epsilon
\end{equation}

\noindent where $\epsilon$ is modeled with a standard Gaussian distribution.
In practice, the compressed input representation $p_{\theta}(\mathbf{z}|\mathbf{v})$ is modeled by two distinct linear layers, each projects input to the same dimensionality with $\mathbf{z}$ for computing $\mu _{\theta}(\mathbf{v})$ and $\Sigma _{\theta}(\mathbf{v})$, where $\mu _{\theta}(\mathbf{v})$ is initialized from the pre-trained weights of the LVLM's vision-language projector, $\Sigma _{\theta}(\mathbf{v})$ is randomly initialized and apply a softplus transform to ensure outputs non-negativity.

\paragraph{Adaptive $\boldsymbol{\beta}$}

Previous studies~\cite{peng2018variational,li2019specializing} have shown that balancing the compression and prediction terms using a Lagrange multiplier $\beta$ is crucial for removing irrelevant information while preserving information that is predictive of the model output.
While setting the $\beta$ to be fixed may prevent the constrained procedure from capturing the dynamic nature of the specific sample.

Inspired by our observation that the smoothness of the similarity distribution between soft visual tokens and the LLM's word embedding strongly correlates with the emergence of object hallucinations, we propose Adaptive Variational Information Bottleneck (\textsc{AdaVIB}). 
The proposal adaptively constrains the injected noise regarding the smoothness of the similarity distribution for capturing the dynamic nature per sample.
Specifically, we use the entropy~\cite{shannon1948mathematical,zhai2023stabilizing,farquhar2024detecting} of the similarity distribution as an indicator to reflect the degree of soft visual tokens suffering from the overconfidence problem.
A sample with a low entropy level refers to a sharp similarity distribution, indicating that the soft visual tokens are prone to be overconfident~\cite{pereyra2017regularizing} when mapping to the LLM's embedding space.
In contrast, a sample with a high entropy level indicates a smooth similarity distribution, corresponding to the uncertainty of the probability distribution~\cite{alemi2018uncertainty}. 
Therefore, ensuring the entropy within a proper value is essential to effectively delivering essential information.
With this intuition, our method computes the smoothing indicator on the fly during the forward propagation in training, relying on the entropy of the similarity distribution per sample, which is formalized as follows:
\begin{equation}
\label{sim_entropy}
H=-\sum_{i=1}^{| V |}{p( E_{LLM}^{( i )} |\mathbf{z} )}\log p( E_{LLM}^{( i )} |\mathbf{z} ) 
\end{equation}
\begin{equation}
p( E_{LLM} |\mathbf{z} ) = \texttt{softmax} ( \mathbf{\bar{z}}\cdot E_{LLM}^{\top} )
\end{equation}

\noindent where $\mathbf{\bar{z}}$ is the average pooling of $\mathbf{z}$ for all soft visual tokens, $E_{LLM}$ is the word embedding of the LLM, $| V |$ is the vocabulary size of the LLM. 
Since entropy $H$ has an unfixed range, we normalize $H$ between 0 and 1 with its maximum value $\log ( | V | )$. 
The adaptive $\beta$ updates as follows:
\begin{equation}
\label{ada_beta}
\beta \gets -\beta \cdot \log ( \frac{H}{\log ( | V | )} ) 
\end{equation}

With this mechanism, a sample with low entropy is trained with high $\beta$, in which the compression term in Equation \ref{eq_vib} dominates the optimization process, thereby constraining the information conveyed by input through adding a significant noise perturbation.
Conversely, a sample with high entropy corresponds to low $\beta$, where the prediction term dominates, preventing the learning process from converging to a worse performance.
Note that the gradient computation for $\beta$ is excluded from the computation graph, thereby the gradient does not flow through adaptive $\beta$.

\section{Experiments}

\subsection{Datasets and Evaluation Metrics}

\paragraph{MSCOCO~\cite{lin2014microsoft}}
The Microsoft Common Objects in Context (MSCOCO) stands as a comprehensive dataset for evaluating various visual tasks, including image recognition, segmentation, and captioning. 
It has more than 300k images with annotations for over 80 object categories.
In this paper, we employ COCO2014 to assess object hallucinations on the image captioning task.
To train the vision-language projector, we randomly select 5000 image-text pairs from LLaVA-150k~\cite{liu2024visual}, which is a set of GPT-generated multi-modal instruction-following data grounded on the images from COCO2014.
Following \citet{zhou2024analyzing}, we additionally select 5000 unique images from the training dataset of COCO2014 to evaluate object hallucinations, ensuring that the selected images do not overlap with those used in training.

We use Caption Hallucination Assessment with Image Relevance (CHAIR)~\cite{rohrbach2018object} metric to evaluate object hallucinations in the image captioning task.
CHAIR assesses object hallucinations by counter objects mentioned in the predicted caption but not present in the grounded image.
It has two commonly used variants, CHAIR$_S$ and CHAIR$_I$, where the former assesses object hallucinations at the sentence level, and the latter assesses at the instance level.

\paragraph{POPE~\cite{li2023evaluating}}

The Polling-based Object Probing Evaluation (POPE) is a widely adopted benchmark for assessing object hallucination on the Visual Question Answering (VQA) task.
We conduct the evaluation on this benchmark to verify the generalization capability of \textsc{AdaVIB} on different tasks.
POPE designs a binary question format, requiring the LVLM to deliver a binary-like answer to discriminate whether the mentioned objects are within the grounded image.
We use the official question sets introduced in \citet{li2023evaluating}.
The dataset comprises three splits:
In \textbf{Random} split, the absent objects are randomly selected from the whole dataset.
In \textbf{Popular} split, the absent objects are chosen from the most frequently appeared objects in the dataset.
In \textbf{Adversarial} split, the absent objects are selected from those frequently co-occurred with ground-truth objects.
Each split is composed of 3000 questions on images taken from the validation set of COCO2014.

We use Accuracy and F1 scores to evaluate model performance, where Accuracy reflects the proportion of samples that correctly predict the golden answer.
F1 score is the harmonic average of precision and recall.
Following~\citet{li2023evaluating}, we use it as the major metric for evaluation.

\subsection{Baselines}

We employ the following competitive baselines to compare with our approach:
\textbf{Chain-of-Thought (CoT)}: \citet{wei2022chain} decomposed a hard problem by generating intermediate steps for the final answer.
Following~\citet{zhou2024analyzing}, we leverage CoT by asking the model to first list the identified objects and then describe the image in terms of these objects.
\textbf{Decoding by Contrasting Layers (DOLA)}: \citet{chuang2023dola} mitigated LLM's hallucinations by contrasting the differences between output logits from the later and earlier layers of LLMs.
\textbf{Teacher}: \citet{saha2024can} integrated several short descriptions into a long-form version. Following~\citet{zhou2024analyzing}, we leverage BLIP-2~\cite{li2023blip} to generate short descriptions as contextual guidance, facilitating a long-form description using the caption generator. 
\textbf{Visual Contrastive Decoding (VCD)}: \citet{leng2024mitigating} alleviated hallucinations by introducing a visual contrastive decoding strategy, effectively decreasing the over-reliance on statistical bias and unimodal priors.
\textbf{LVLM Hallucination Revisor (LURE)}: ~\citet{zhou2024analyzing} introduced a post-hoc rectification method to train a hallucination revisor for rectifying initially generated descriptions.

Apart from the above baselines, we employ two distinct LVLMs as backbone frameworks, i.e., MiniGPT-4 and LLaVa-1.5.
We employ both original and fine-tuned (FT) version as baselines.
Additionally, we investigate the effectiveness of the variant without adaptive $\beta$ (w/o Ada $\beta$) in Equation~\ref{ada_beta}, and the one without reparameterization strategy (w/o Repara.) in Equation~\ref{repara_trick} for \textsc{AdaVIB}.

\subsection{Implementation Details}

We employ Vicuna-7B as the caption generator for both MiniGPT4 and LLaVa-1.5.
We train all models in one epoch to avoid overfitting. 
All hyperparameters of baselines are selected via cross-validation on the training dataset of MSCOCO.
Specifically, the Lagrange multiplier $\beta$ is set to $\beta =1e^{-7}$ unless explicitly specified.
We set the batch size to 2 with gradient accumulation steps to 8.
The maximum sequence length during training is set to 512.
We use greedy decoding with a maximum decoding length of 256 during inference.
We set the learning rate to $3e^{-5}$ with a weight decay of 0.05, and use a linear warm-up schedule for the first 1/10 optimization steps, followed by a polynomial decay.
We use an A100-PCIE-40G GPU for training, which takes approximately 20 minutes for MiniGPT4 and 40 minutes for LLaVa-1.5.
There are around 6.3M and 25.2M trainable parameters for MiniGPT4 and LLaVa-1.5, respectively.

\subsection{Evaluation Results}

\begin{table}[t]
\centering
\setlength{\tabcolsep}{8pt}
\resizebox{0.8\linewidth}{!}{
\begin{tabular}{l | cc | cc}
    \hline
    \hline

    \multirow{2}*{Model} & 
    \multicolumn{2}{c|}{\emph{MiniGPT-4}} & 
    \multicolumn{2}{c}{\emph{LLaVa}} \\
    
    ~ & $C_S\downarrow$ & $C_I\downarrow$ & $C_S\downarrow$ & $C_I\downarrow$ \\

    \hline
    Teacher{\scriptsize{13B}} & 24.0 & 5.7 & 49.9 & 9.3 \\
    CoT{\scriptsize{13B}} & 31.6 & 9.4 & 47.6  & 9.0 \\
    DOLA{\scriptsize{7B}}$^{\dagger}$ & 26.7 & 7.9 & 31.9 & 8.3 \\
    VCD{\scriptsize{7B}}$^{\dagger}$ & 24.4 & 7.5 & 26.2 & 7.8 \\
    LURE{\scriptsize{13B}} & 19.7 & \textbf{4.9} & 27.1 & 6.4 \\

    \hline
    Original{\scriptsize{7B}} & 27.1 & 7.8 & 47.8 & 10.8 \\
    
    FT{\scriptsize{7B}} & 19.3 & 6.4 & 26.7 & 7.2 \\
    
    \textbf{\textsc{AdaVIB}{\scriptsize{7B}}} & \textbf{16.2} & \underline{5.2} & \textbf{18.4} & \textbf{5.5} \\
    
    ~ w/o Ada $\beta$ & \underline{17.1} & 5.6 & \underline{19.2} & \underline{5.7} \\
    
    ~ w/o Repara.  & 18.5 & 6.2 & 22.3 & 6.4 \\
    
    \hline
    \hline
\end{tabular}
}
\caption{Experimental results on MSCOCO. The evaluation is performed using CHAIR$_S$ ($C_S$) and CHAIR$_I$ ($C_I$), with smaller values indicating fewer object hallucinations. $^{\dagger}$ means rerun using their released code. The best and the second-best are marked in bold and underlined, respectively.}
\label{coco_main_results}
\end{table}

\paragraph{Results on MSCOCO}
Table \ref{coco_main_results} reports the results on MSCOCO, we have following observations: 
First, the proposed \textsc{AdaVIB} yields consistent improvements across both MiniGPT4 and LLaVa-1.5 model architectures.
Specifically, it outperforms one of the strongest baseline LURE{\scriptsize{13B}} by around 14.5\% under both CHAIR$_S$ and CHAIR$_I$ metrics, especially under the CHAIR$_S$ metric, it surpasses LURE by around 24.9\% across two distinct model architectures.
Second, \textsc{AdaVIB} substantially outperforms the vanilla fine-tuning method.
The improvement of \textsc{AdaVIB} over fine-tuning under CHAIR$_S$ and CHAIR$_I$ metrics are 23.6\% and 21.2\%, respectively.
This indicates that the vanilla fully connected layer may not be strong enough to model the sophisticated mapping relations between vision and language. 
In contrast, our approach can improve the vision-language alignment by better compressing irrelevant visual information while maintaining relevant information for generating hallucination-free descriptions.
Third, the ablation results indicate the effectiveness of each component.
Specifically, by removing adaptive $\beta$ in equation \ref{ada_beta}, the results drop over 5.0\% on average across two model architectures.
Notably, by removing the reparameterization strategy in equation \ref{repara_trick}, the results drop over 15.0\% on average, indicating the significance of this component in mitigating object hallucinations on the MSCOCO dataset.

\paragraph{Results on POPE}

Table ~\ref{pope_main_results} reports the results on POPE. 
We observe that the improvements of \textsc{AdaVIB} on Popular are clearer than on Random and Adversarial.
Concretely, compared with one of the strongest baselines LURE on MiniGPT4 backbone, the improvements of \textsc{AdaVIB} on Popular (8.1\%) are more significant than on Random (4.8\%) and Adversarial (3.9\%).
Similar observations can be found when using the LLaVa backbone, \textsc{AdaVIB} achieves 7.5\% improvements on Popular, while 2.5\% improvements on Random and Adversarial by average.
These observations indicate the advantages of \textsc{AdaVIB} in effectively mitigating the statistical bias arising from the most frequently appeared objects during training, reducing the over-reliance on irrelevant features, thereby alleviating object hallucinations.
Additionally, the ablation results on POPE indicate the effectiveness of \textsc{AdaVIB}.
Concretely, removing adaptive $\beta$ decreases \textsc{AdaVIB} by 1.5\% and 1.4\% on MiniGPT4 and LLaVA-1.5 by average, respectively.
While removing the reparameterization strategy results in performance degradation by 2.1\% and 2.2\% on MiniGPT4 and LLaVA-1.5 by average, respectively.
This observation indicates that both adaptive $\beta$ and reparameterization strategy is important to improve the performance of \textsc{AdaVIB} on the POPE dataset.

\begin{table}[t]
\centering
\setlength{\tabcolsep}{2.5pt}
\resizebox{0.9\linewidth}{!}{
\begin{tabular}{l | cc | cc | cc}
    \hline
    \hline

    \multirow{2}*{Model} & 
    \multicolumn{2}{c|}{Random} & 
    \multicolumn{2}{c|}{Popular} & 
    \multicolumn{2}{c}{Adversarial} \\
    
    ~ & $ACC\uparrow$ & $F1\uparrow$ & $ACC\uparrow$ & $F1\uparrow$ & $ACC\uparrow$ & $F1\uparrow$ \\

    \hline
    
    \multicolumn{7}{c}{\emph{MiniGPT4}} \\

    \hline

    DOLA{\scriptsize{7B}}$^{\dagger}$ & 76.6 & 77.9 & 65.8 & 69.3 & 63.3 & 68.5 \\
    
    VCD{\scriptsize{7B}}$^{\dagger}$ & 78.4 & 80.2 & 68.5 & 72.0 & 64.4 & 70.3 \\

    LURE{\scriptsize{13B}}$^{\dagger}$ & 78.0 & 79.2 & 66.0 & 70.3 & 63.9 & 71.1 \\

    FT{\scriptsize{7B}} & 79.0 & 80.8 & \underline{70.0} & 72.9 & 65.0 & 71.2 \\

    \textbf{\textsc{AdaVIB}{\scriptsize{7B}}} & \textbf{81.2} & \textbf{83.6} & \textbf{71.1} & \textbf{76.3} & \textbf{66.7} & \textbf{73.6} \\
    
    ~ w/o Ada $\beta$ & \underline{80.0} & \underline{82.2} & 69.8 & \underline{75.6} & 65.3 & \underline{73.1} \\
    
    ~ w/o Repara. & 79.4 & 82.0 & 69.5 & 74.0 & \underline{65.6} & 72.5 \\ 

    \hline
    
    \multicolumn{7}{c}{\emph{LLaVa}} \\

    \hline

    DOLA{\scriptsize{7B}}$^{\dagger}$ & 84.0 & 84.4 & 79.5 & 82.2 & 76.6 & 77.2 \\
    
    VCD{\scriptsize{7B}}$^{\dagger}$ & \textbf{86.8} & \underline{86.2} & 83.2 & 84.3 & \textbf{80.1} & \underline{79.6} \\
    
    LURE{\scriptsize{13B}} & \underline{86.3} & 85.8 & 80.3 & 80.7 & 77.2 & 78.4 \\

    FT{\scriptsize{7B}} & 84.3 & 85.8 & 84.9 & 84.0 & 76.5 & 77.7 \\
    
    \textbf{\textsc{AdaVIB}{\scriptsize{7B}}} & \underline{86.3} & \textbf{87.0} & \textbf{86.3} & \textbf{86.8} & \underline{78.0} & \textbf{80.4} \\
    
    ~ w/o Ada $\beta$ & 85.4 & 86.0 & \underline{85.0} & \underline{85.6} & 76.9 & 79.0 \\
    
    ~ w/o Repara. & 84.9 & 85.3 & 84.5 & 85.1 & 76.2 & 78.1 \\
    
    \hline
    \hline
\end{tabular}
}
\caption{Experimental results on POPE. Larger values indicate less hallucinations.}
\label{pope_main_results}
\end{table}

\subsection{Discussions}

\begin{figure}[t]
\centering
\includegraphics[width=6.5cm]{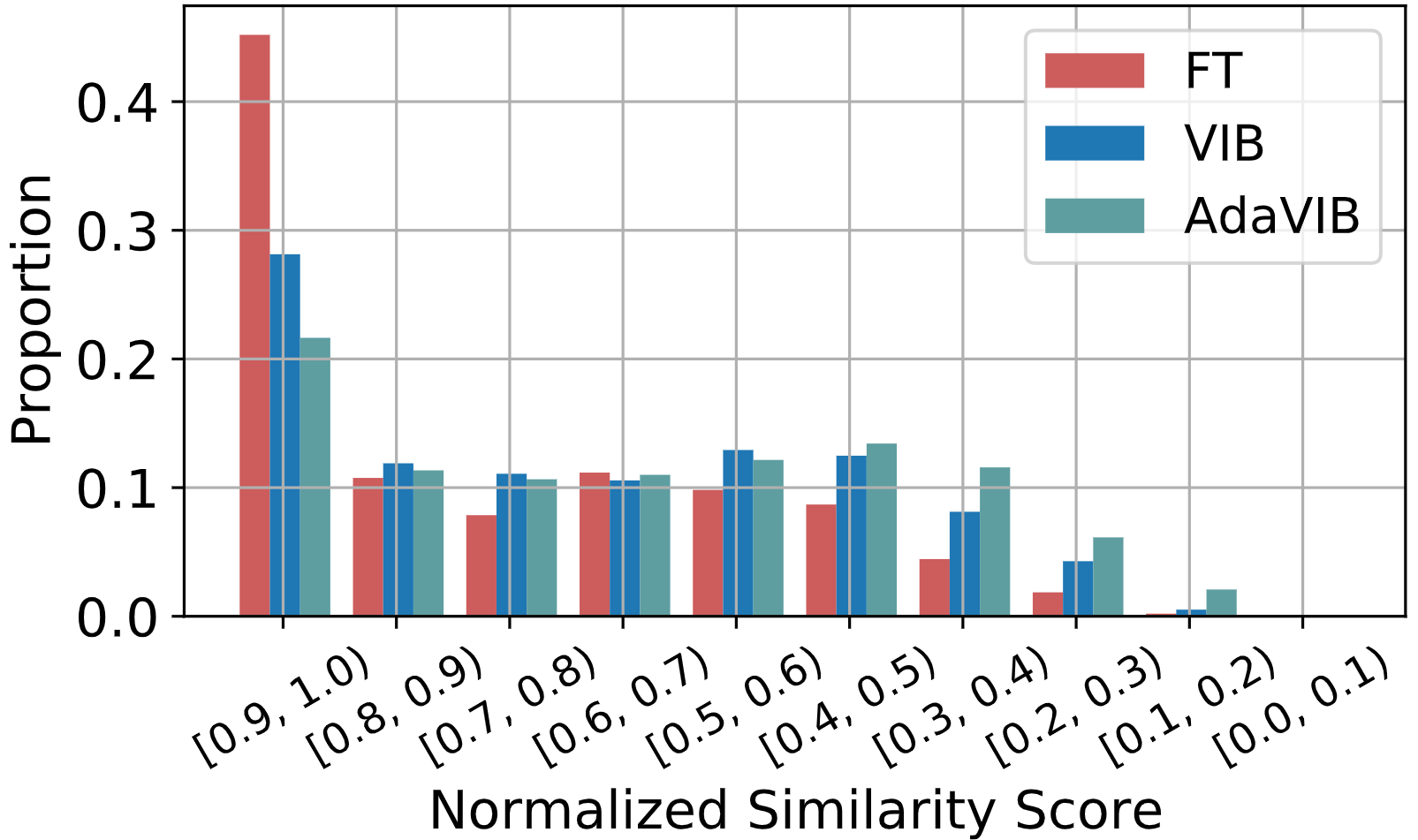}
\caption{Distribution of the max similarity score between soft visual tokens and LLM's word embedding. We use MiniGPT4 as the model backbone. The x-axis denotes the normalized similarity score, ranging from 0-1. The y-axis denotes the proportion of hallucinated samples in a specific range to overall hallucinated samples.}
\label{overconf_ana}
\end{figure}

\paragraph{Does the \textsc{AdaVIB} mitigate object hallucinations by alleviating the \emph{overconfidence} problem?}
To address this, we figure out the proportion of hallucinated samples in a specific range of the max similarity score to overall hallucinated samples.
Figure \ref{overconf_ana} presents the results that employ the MiniGPT4 as the model backbone on the MSCOCO dataset. We have the following observations:
First, for the fine-tuned (FT) variant, the similarity score ranging $[0.9, 1.0)$ dominates the hallucinated samples, with an average similarity entropy of 0.64.
A low entropy denotes a sharp similarity distribution, indicating that the fine-tuned variant is prone to overconfidence when visual tokens map to the LLM's word embedding space.
Second, both VIB and \textsc{AdaVIB} have a smoother distribution than the fine-tuned variant, with a consistent improvement on alleviating object hallucinations from the results reported in Table \ref{coco_main_results} and Table \ref{pope_main_results}. 
This observation indicates that mitigating the \emph{overconfidence} problem is one of the effective solutions to mitigate object hallucinations.
Third, \textsc{AdaVIB} has fewer hallucinated samples with a large max similarity score than VIB.
It has a smoother distribution than VIB, with a larger average similarity entropy (1.30 v.s. 0.97).
This result indicates that the proposed adaptive $\beta$ mitigates object hallucinations by effectively controlling the smoothness of the similarity distribution.

\paragraph{Does the \textsc{AdaVIB} alleviate the \emph{overconfidence} issue by effectively constraining irrelevant information?}

\begin{figure}[t]
\centering
\includegraphics[width=7.cm]{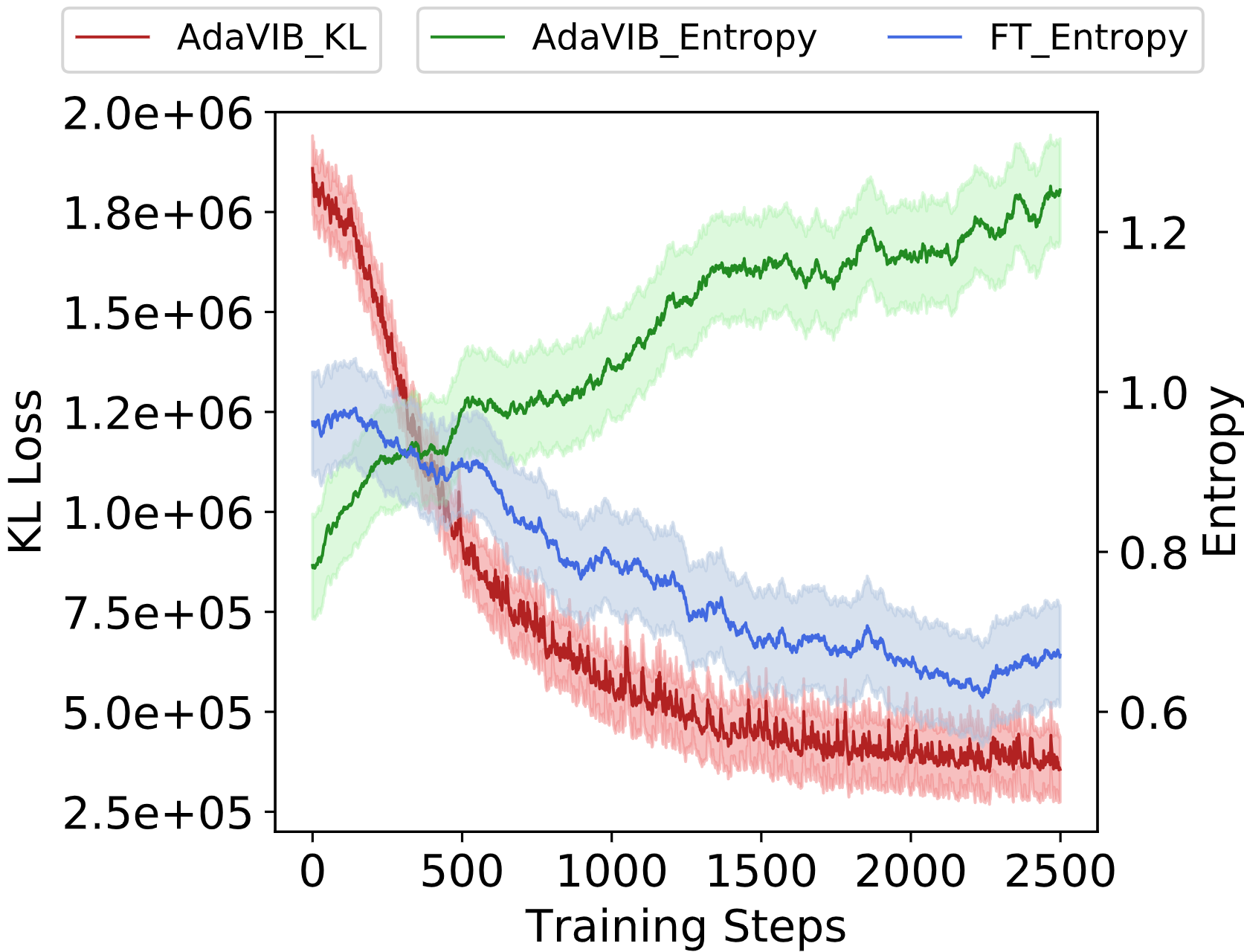}
\caption{Correlation between the KL loss (Equation \ref{kl_loss}) and the similarity entropy (Equation \ref{sim_entropy}) over the course of training. All curves are smoothed by exponential moving average for better understanding the tendency.}
\label{learning_curves}
\end{figure}

In addressing this question, we present the learning curves of the KL loss for \textsc{AdaVIB}, as well as the curves of similarity entropy for both \textsc{AdaVIB} and fine-tuning.
From Figure \ref{learning_curves}, we have two main observations:
First, with the training progress, the KL loss of \textsc{AdaVIB} (red curve) is decreased, accompanied by the increase of its similarity entropy (green curve).
This phenomenon indicates that the compression term in Equation \ref{eq_vib} effectively compresses the irrelevant information by adding stochastic noise, thereby progressively reducing the overconfidence in irrelevant visual features between the visual tokens and the LLM's word embedding. 
Second, compared with the curves of similarity entropy between \textsc{AdaVIB} and fine-tuning, the entropy curve of the fine-tuning variant severely decreases with the training process going on, indicating the emergence of the overconfidence problem.
In contrast, \textsc{AdaVIB} improves the performance by progressively smoothing the similarity distribution, avoiding overconfidence in irrelevant visual features.

\paragraph{How does different $\beta$ impact the model performance on mitigating object hallucinations?}

\begin{figure}[t]
\centering
    \subfigure[MiniGPT4]
    {
    \begin{minipage}{3.9cm}
    \label{minigpt4_impact}
    \centering
    \includegraphics[width=4.1cm]{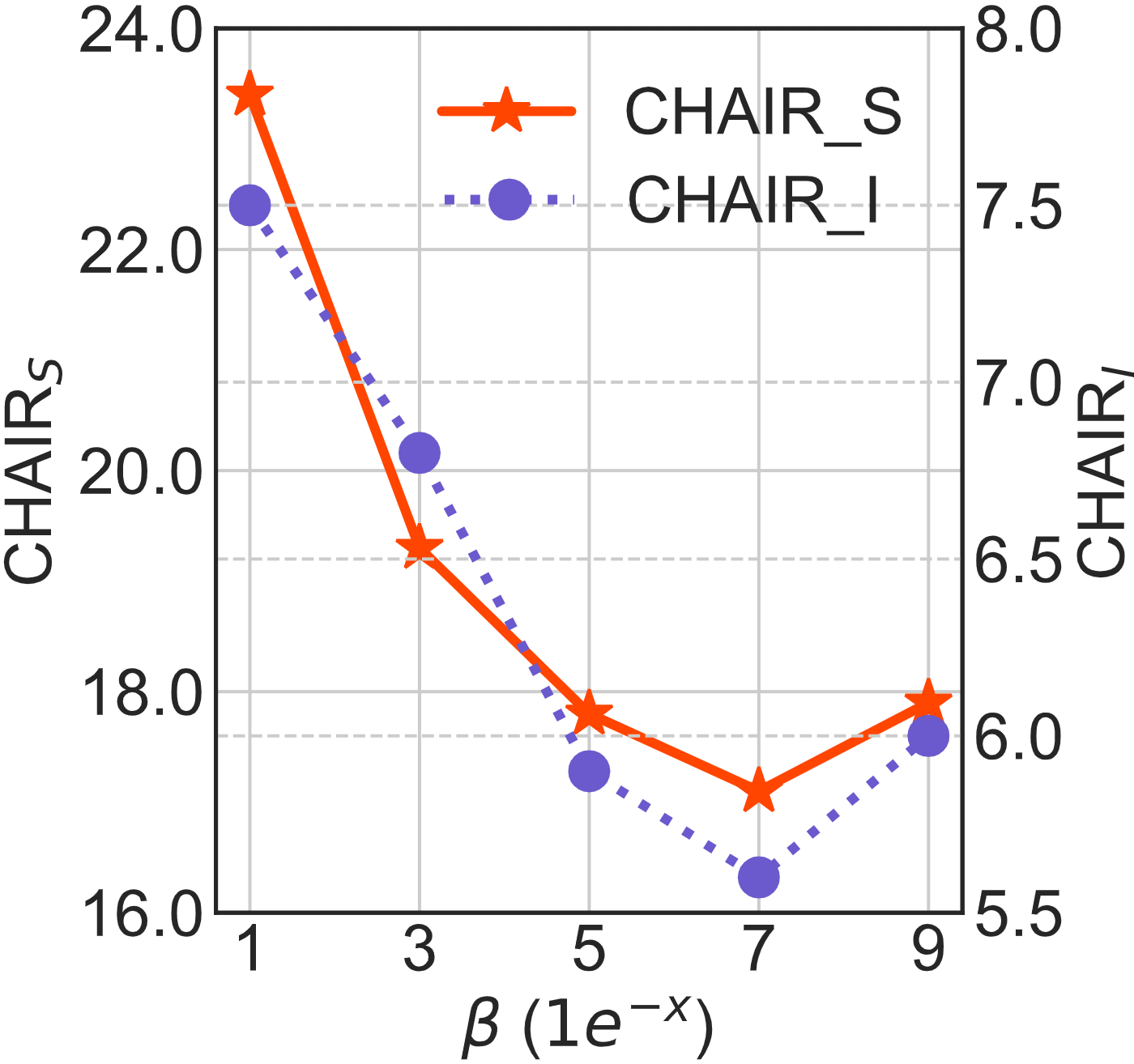}
    \end{minipage}
    }
    \subfigure[LLaVa-1.5]
    {
    \begin{minipage}{3.9cm}
    \label{llava_impact}
    \centering
    \includegraphics[width=4.1cm]{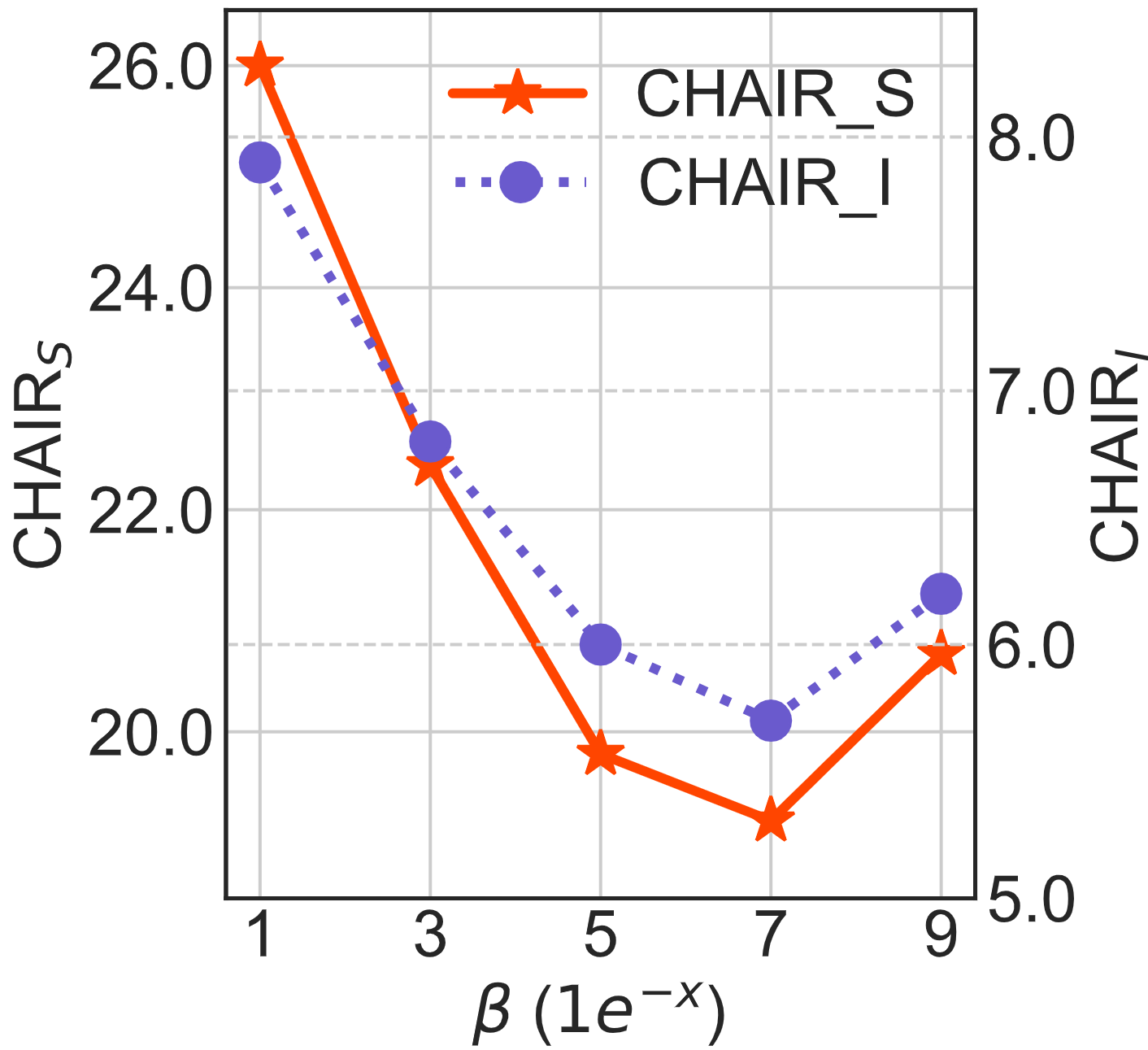}
    \end{minipage}
    }
\caption{Impact of different $\beta$ on object hallucinations. Figure \ref{minigpt4_impact} and Figure \ref{llava_impact} present the results that leverages MiniGPT4 and LLaVa-1.5 as backbone, respectively.} 
\label{beta_impact}
\end{figure}

To answer this question, we employ both MiniGPT4 and LLaVa-1.5 as the backbone, adjusting $\beta$ and analyzing the change of the CHAIR score on MSCOCO evaluation dataset.
From Figure \ref{beta_impact}, we observe that the performance of both CHAIR$_S$ and CHAIR$_I$ score improves along with the reduction of $\beta$ from $\beta =1e^{-1}$ to $\beta =1e^{-7}$, followed by an obvious performance degradation with $\beta =1e^{-9}$.
This observation indicates that setting the $\beta$ to a proper value is essential to effectively constrain the information flow while preserving the useful information as much as possible. 
A large $\beta$ introduces a big noise during the model optimization, resulting in the model scarcely capturing the essentials relevant to the golden answer, thereby converging to a worse performance.  
When the $\beta$ is too small, the model tends to preserve relevant information rather than compress the irrelevant features present in the original input, delivering a sub-optimal performance in mitigating object hallucinations.
Moreover, setting $\beta$ to a fixed value, regardless of the dynamic nature per sample, cannot achieve the optimal performance in mitigating hallucinations.
This conclusion is consistent with the results reported in Table \ref{coco_main_results}, where removing the adaptive $\beta$ results in a worse performance compared to the variant with the adaptive one.

\paragraph{How does \textsc{AdaVIB} perform compared to other regularization strategies?}
In addition to the fine-tuning baseline that applies the weight decay~\cite{loshchilov2017decoupled} as a regularizer.
We also deploy dropout~\cite{srivastava2014dropout} to the input (DRP$_{in}$) and output (DRP$_{out}$) of the vision-language projector upon the fine-tuning method, with the dropout rate of 0.1.
Table \ref{dropout_ana} presents results. 
We observe that variants equipped with dropout effectively reduce object hallucinations compared to the FT, while they still have a performance gap between \textsc{AdaVIB}.
This indicates the superiority of \textsc{AdaVIB} in the trade-off between compressing irrelevant while preserving relevant information to represent visual tokens precisely.
\begin{table}[h]
\centering
\setlength{\tabcolsep}{8pt}
\resizebox{0.75\linewidth}{!}{
\begin{tabular}{l | cc | cc}
    \hline
    \hline

    \multirow{2}*{Model} & 
    \multicolumn{2}{c|}{\emph{MiniGPT-4}} & 
    \multicolumn{2}{c}{\emph{LLaVa}} \\
    
    ~ & $C_S\downarrow$ & $C_I\downarrow$ & $C_S\downarrow$ & $C_I\downarrow$ \\

    \hline
    
    FT & 19.3 & 6.4 & 26.7 & 7.2 \\
    
    FT+DRP$_{in}$ & 18.5 & 6.1 & 23.4 & 6.4 \\

    FT+DRP$_{out}$ & 18.6 & 5.9 & 22.7 & 6.2 \\
    
    \textbf{\textsc{AdaVIB}} & \textbf{16.2} & \textbf{5.2} & \textbf{18.4} & \textbf{5.5} \\

    \hline
    \hline
\end{tabular}
}
\caption{Impact on different regularization strategies. ``DRP'' is the abbreviation of the ``dropout''.}
\label{dropout_ana}
\end{table}
\section{Conclusions}

In this paper, we propose using VIB to mitigate object hallucinations.
The proposal is based on our observation that the object hallucination can be attributed to the overconfidence in irrelevant visual features when soft visual tokens project to the word embedding space of LLM.
Motivated by our observation, we propose \textsc{AdaVIB} to adaptively constrain irrelevant information regarding the smoothness of similarity distribution.
Experimental results and comprehensive analysis demonstrate the effectiveness of our approach with consistent improvements over competitive baselines.

\section{Acknowledgments}

This work was supported by the National Natural Science Foundation of China (No. 62372126, 62372129, U2436208, 62272119, 62276017, 62072130), the Guangdong Basic and Applied Basic Research Foundation (No. 2023A1515030142), the Key Technologies R\&D Program of Guangdong Province (No. 2024B0101010002), the Strategic Research and Consulting Project of the Chinese Academy of Engineering (No. 2023-JB-13), and the State Key Laboratory of Complex \& Critical Software Environment (No. SKLCCSE-2024ZX-18).

\bibliography{aaai25}

\end{document}